\documentclass[conference]{IEEEtran}
\IEEEoverridecommandlockouts
\usepackage{cite}
\usepackage{amsmath,amssymb,amsfonts}
\usepackage{algorithmic}
\usepackage{graphicx}
\usepackage{textcomp}
\usepackage{xcolor}
\usepackage{cleveref}
\usepackage{multirow}
\usepackage{array}
\usepackage{booktabs}
\usepackage{xspace}
\def\BibTeX{{\rm B\kern-.05em{\sc i\kern-.025em b}\kern-.08em
    T\kern-.1667em\lower.7ex\hbox{E}\kern-.125emX}}
\begin{document}

\title{UniVid: Pyramid Diffusion Model for High Quality Video Generation}


\author{Xinyu Xiao, Binbin Yang, Tingtian Li, Yipeng Yu, Sen Lei}

\maketitle

\begin{abstract}
Diffusion-based text-to-video generation (T2V) or image-to-video (I2V) generation have emerged as a prominent research focus. However, there exists a challenge in integrating the two generative paradigms into a unified model. In this paper, we present a unified video generation model (\emph{Univid}) with hybrid conditions of the text prompt and reference image. Given these two available controls, our model can extract objects' appearance and their motion descriptions from textual prompts, while obtaining texture details and structural information from image clues to guide the video generation process. Specifically, we scale up the pre-trained text-to-image diffusion model for generating temporally coherent frames via introducing our temporal-pyramid cross-frame spatial-temporal attention modules and  convolutions. To support bimodal control, we introduce a dual-stream cross-attention mechanism, whose attention scores can be freely re-weighted for interpolation of between single and two modalities controls during inference. Extensive experiments showcase that our \emph{Univid} achieves superior temporal coherence on T2V, I2V and (T+I)2V tasks.
\end{abstract}


\section{Introduction}
\label{sec:intro}




In recent studies~\cite{ho2022imagen, singer2022make, ge2023preserve, blattmann2023align}, researchers have endeavored to replicate such success in video generation by training a text-to-video~(T2V) diffusion model. Text prompts are friendly interface for image synthesis, but for videos, using a reference image as a prompt can be even handier for users. In this work, we seek to develop a unified video generation model that harnesses the power of both text and image prompts, granting users the flexibility to render a video using images, text, or even a combination of both modalities as the control, \emph{i.e.}, T2V, I2V and (T+I)2V.

\begin{figure}[t!]
\begin{center}
\includegraphics[width=1.0\linewidth]{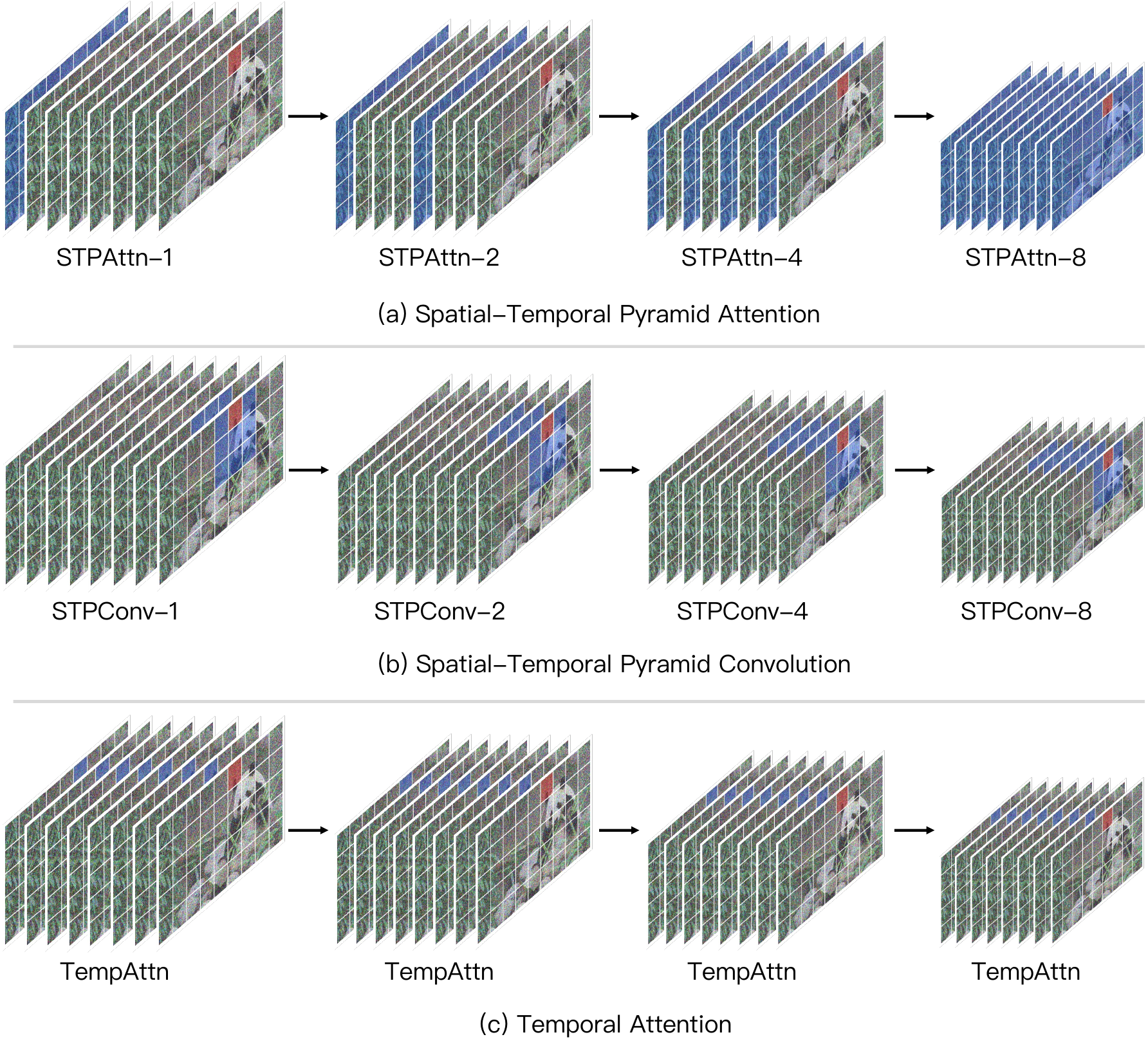}
\end{center}
\caption{An illustration of our proposed pyramid spatial-temporal attention ($\mathtt{PSTAttn}\emph{-f}$), pyramid spatial-temporal convolution ($\mathtt{PSTConv}\emph{-f}$) and temporal attention, where $\emph{f} \in \{1,2,4,8\}$ indicates the downsampling factors in the U-Net layers. The red patches in frames denotes the queries of attention, and the blue ones denotes the keys / values of attention. (We also visualize the receptive field of $\mathtt{PSTConv}\emph{-f}$ in a similar manner.)}
\label{fig:pyramid_fig}
\end{figure}





Most recent works~\cite{blattmann2023align, singer2022make} turned the T2I diffusion model into a video generator by inserting a $1D$ temporal layer after each spatial module or turning spatial attention to full attention.  Based on the U-Net backbone in Stable Diffusion~\cite{rombach2022high}, in this work, we introduce a pyramid spatial-temporal attention module to improve dynamics modeling and capture motion clues by exploring inter-frame region relationships with dynamic visual tempo.
As illustrated in \cref{fig:pyramid_fig}, $3D$ spatial-temporal attention can fully model the motion clue of the moving panda by learning the correlations of the panda in different frames while $1D$ temporal attention falls short because only per-pixel temporal dependency is considered. 




To this end, we present $\emph{Univid}$, a unified video diffusion model which generates high quality videos by using a flexible combination of descriptive sentences and reference images as hybrid controls. 
For the temporal extension, we draw inspiration from the success of modeling motion dynamics with multi-level visual tempo in video understanding~\cite{yang2020temporal, zhang2022actionformer}, and propose to incorporate pyramid spatial-temporal attention and convolution modules into the basic blocks of our diffusion U-Net. 
By adapting spatial pyramid architecture~~\cite{harada2011discriminative, he2015spatial, bosch2007representing} to a temporal dimension, our pyramid spatial-temporal attention progressively learns the motion clues across frames in a temporally coarse-to-fine manner.
Specifically, in the middle blocks of the U-Net, we perform dense spatial-temporal semantic alignments among the high-level frame features while sparse alignments in external blocks.

To effectively encode appearance details from images and semantic descriptions from text for guidance in generation, we propose extending the regular cross-attention module in Stable Diffusion into a dual-stream cross-attention module.
Similar to the text branch, we encode the reference image using CLIP~\cite{radford2021learning} image encoder and perform the visual cross-attention between the input feature and the sequence of visual patch embeddings. The visual and textual cross-attention operate in a two-path manner, and we fuse their results as the output feature. 





To sum up, our main contributions are listed as follows:
\begin{itemize}
\item We develop a high-quality video generation model, namely \emph{UniVid} which can receives the flexible combinational controls from text prompts and reference images.

\item We propose to empower the pre-trained text-to-image diffusion model with the ability of temporal motion modeling by incorporating pyramid spatial-temporal attention and convolution modules into the 2D image generation backbone.

\item We propose a dual-stream cross-attention for aligning the semantics of the video features and the user-provided text prompts and reference images, in a two-path manner.

\item Zero-shot generative evaluations on UCF-101 and MSR-VTT demonstrates that \emph{UniVId} achieves better generative performance than existing methods.

\end{itemize}

\section{Preliminaries}
\label{sec:preliminaries}

\begin{figure*}[ht!]
\begin{center}
\includegraphics[width=1.0\linewidth]{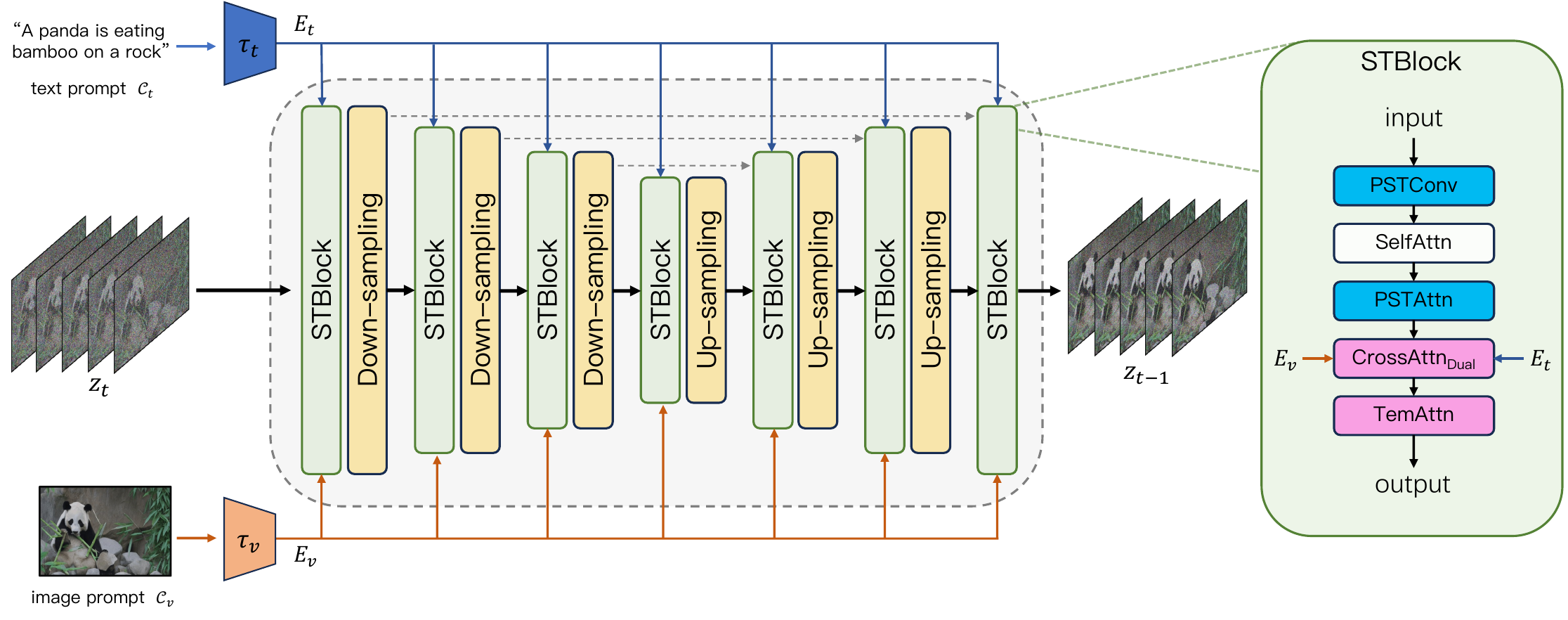}
\end{center}
\caption{An overview of UniVid. For the latent $z_t$, a text prompt $\mathcal{C}_t$ and an image prompt $\mathcal{C}_v$ are used as hybrid condition to guide the denoising (generation) process. Specifically, we use a CLIP text encoder $\tau_t$ and an CLIP image encoder $\tau_v$ to obtain the linguistic token embedding $E_t$ and visual patch embedding $E_v$, respectively. The U-Net architecture of UniVid is built by stacking multiple basic STBlocks, with each block consisting of pyramid spatial-temporal convolution~($\mathtt{PSTConv}$), self-attention~($\mathtt{SelfAttn}$), pyramid spatial-temporal attention~($\mathtt{PSTAttn}$), dual-stream cross-attention~($\mathtt{CrossAttn}$) and temporal attention~($\mathtt{TemAttn}$). Best viewed in color.}
\label{fig:overview}
\end{figure*}

Diffusion model~\cite{ho2020denoising, rombach2022high, ramesh2022hierarchical, saharia2022photorealistic} is a class of likelihood-based generative models, consisting of a diffusion process and a denoising process. 
Formally, a data sample $x_0 \sim q(x_0)$ is progressively blurred by a sequence of Gaussian noises:
\begin{equation}
    q(x_t | x_{t-1}) = \mathcal{N}(x_t; \sqrt{\alpha_t}x_{t-1}, (1-\alpha_t) \mathbf{I}),
\end{equation}
where $\{\alpha_t\}_{t=1}^T$ are the noise magnitudes. According to the properties of Markov chain and Gaussian noise, $x_T$ will approximately follow a standard Gaussian distribution $\mathcal{N}(\mathbf{0}, \mathbf{I})$ if the chain length $T$ is large.
The generative process is then defined by reversely denoising from a standard Gaussian noise, \emph{i.e.}, $x_T \sim \mathcal{N}(\mathbf{0}, \mathbf{I})$. Denoising diffusion probabilistic models~\cite{ho2020denoising} revealed that training a diffusion model can be reduced to learning to estimate the noise $\epsilon$ added to $x_0$ for each timestep $t$:
\begin{equation}
\label{eq:loss_dm}
    \mathcal{L}_{\text{DM}}(\theta) = \mathbb{E}_{t, x_0, \epsilon\sim\mathcal{N}(\mathbf{0}, \mathbf{I})}\left[\Vert\epsilon_{\theta}(x_t, t) - \epsilon\Vert^2\right],
\end{equation}
where $\epsilon_{\theta}(x_t, t)$ is a noise predictor implemented as a U-Net~\cite{ronneberger2015u} with parameters $\theta$.

Latent Diffusion Model~\cite{rombach2022high} is a variant of Diffusion Models, which performs the diffusion process in the compressed latent space of VAE~\cite{kingma2013auto} rather than the original pixel space. Taking Stable Diffusion~\cite{rombach2022high} as an example, the training loss for the text-to-image latent diffusion model can be formulated as:
\begin{equation}
\label{eq:loss_ldm}
\small
    \mathcal{L}_{\text{LDM}}(\theta) = \mathbb{E}_{t, z_0, \epsilon\sim\mathcal{N}(\mathbf{0}, \mathbf{I})}\left[\Vert\epsilon_{\theta}(z_t, \mathcal{C}_t, t) - \epsilon\Vert^2\right],
\end{equation}
where $z_0 = \mathcal{E}(x_0)$ is the encoded latent for $x_0$, $z_t$ is its noisy version in timestep $t$, $\epsilon_{\theta}(z_t, \mathcal{C}_t, t)$ is a noise predictor with text condition $\mathcal{C}_t$.

\section{Method}
\label{sec:method}

In this section, we will present how we propose a cascaded latent diffusion model, \emph{i.e.}, \emph{Univid}, for controllable high-quality video generation, which allows for dual-stream control from both textual descriptions and reference images. The overall framework of \emph{UidVid} is illustrated in \cref{fig:overview}. The key module of \emph{UidVid} is a base latent video diffusion model, which learns to generate a frame sequences according to the input text and optional image prompts.

\subsection{Video Generation with Dual-Stream Controls}
\label{sec:base_model}




In this paper, we aim to leverage the off-the-shelf Text-to-Image~(T2I) Latent Diffusion Model, \emph{e.g.}, Stable Diffusion~\cite{rombach2022high} as a content prior and extend it into a flexible video generator that can be jointly controlled by textual prompts and reference images. The U-Net architecture in Stable Diffusion is built by stacking multiple basic blocks, with each block consisting of 2D Convolution, Self-Attention and Cross-Attention modules:
\begin{equation}
\label{eq:sd_block}
\mathtt{SD Block} = \mathtt{Conv2D} \circ \mathtt{SelfAttn} \circ \mathtt{CrossAttn}.
\end{equation}
The $\mathtt{SDBlock}$ updates features by leveraging the pixel correlations within each frame and establishing alignments between pixels and input texts. However, it is not directly applicable for generating consecutive frames with motion-aware content correlation, as it lacks the necessary spatial-temporal modeling capability.
Hence, our primary objective is to extend the spatial $\mathtt{SD Block}$ to a spatial-temporal version, namely $\mathtt{ST Block}$ that can capture temporal information, thereby enabling cross-frame semantic alignment and supporting dual-modal input control. The formulation of our $\mathtt{ST Blcok}$ can be described as follows:
\begin{align}
\label{eq:st_block}
\mathtt{ST Block} = &\;\mathtt{\textcolor{cyan}{PSTConv}} \circ \mathtt{SelfAttn} \circ \mathtt{\textcolor{cyan}{PSTAttn}} \;\circ \nonumber \\
  &\; \mathtt{\textcolor{magenta}{CrossAttn_{Dual}}} \circ \mathtt{\textcolor{magenta}{TemAttn}},
\end{align}
where $\mathtt{SelfAttn}$ is the self-attention module, $\mathtt{\textcolor{cyan}{PSTConv}}$ and $\mathtt{\textcolor{cyan}{PSTAttn}}$ are pyramid spatial-temporal convolution and attention, $\mathtt{\textcolor{magenta}{CrossAttn_{Dual}}}$ extends the $\mathtt{CrossAttn}$ in Stable Diffusion for dual-stream guidance of both text and image prompts and $\mathtt{\textcolor{magenta}{TemAttn}}$ is the temporal attention layer.



\subsubsection{Pyramid Spatial-Temporal Modules}
The $\mathtt{SelfAttn}$ and $\mathtt{Conv2D}$ in $\mathtt{SDBlock}$ independently update the spatial features within each frame without explicitly considering their semantic correlations, which often leads to appearance inconsistencies across different frames. To enhance the cross-frame consistency and incorporate motion modeling, we introduce a spatial-temporal attention module, namely $\mathtt{\textcolor{cyan}{PSTAttn}}$ and extend the $\mathtt{Conv2D}$ modules to $\mathtt{\textcolor{cyan}{PSTConv}}$ via stacking a $1D$ temporal convolution after each $2D$ convolution.



Taking inspiration from the success of modeling motion dynamics with multi-level visual tempo in video understanding~\cite{yang2020temporal, feichtenhofer2019slowfast, zhou2018temporal, zhang2022actionformer}, we propose to perform a dynamic tempo modulation in our $\mathtt{\textcolor{cyan}{PSTAttn}}$ and $\mathtt{\textcolor{cyan}{PSTConv}}$ for capturing the complex motion structures inside videos. By adapting the spirit of spatial-pyramid~\cite{harada2011discriminative, he2015spatial, bosch2007representing} to our temporal structures, our pyramid spatial-temporal modules progressively learns the motion clues across frames in a temporally coarse-to-fine manner. As illustrated in \cref{fig:pyramid_fig}, we achieve dense cross-frame semantic correlations in the middle blocks~(with lower resolutions) of the U-Net while sparse correlations in external layers.
Specifically, note that the encoded latent of the input video, $z = \mathcal{E}(x) \in \mathbb{R}^{F\times C\times H\times W}$ will be downsampled in the U-Net using factors $\emph{f} \in \{1,2,4,8\}$.\footnote{Here we use $\emph{f} = 1$ to denotes the features before the first downsampling operation or after the last upsampling operation.} We refer to the modules that process the downsampled features with resolution $\frac{H}{\emph{f}}\times \frac{W}{\emph{f}}$ as $\mathtt{\textcolor{cyan}{PSTAttn}}$\textcolor{cyan}{\emph{-f}} and $\mathtt{\textcolor{cyan}{PSTConv}}$\textcolor{cyan}{\emph{-f}} (a larger $\emph{f}$ indicates the module is closer to the middle layer). In $\mathtt{\textcolor{cyan}{PSTAttn}}$\textcolor{cyan}{\emph{-f}}, each frame selectively attends to a sub-sequence of frames sampled with a dynamic temporal step-size $r_{\emph{f}}$, by exploring the pixel-to-pixel correlations across frames. To achieve this, we evenly divide the temporal interval $[1,2,...,F]$ into $n_\emph{f} = \frac{F}{r_{\emph{f}}}$ segments, and then sample $n_\emph{f}$ reference frames from each segment to form the reference-frame set $\Gamma$.
Formally, it can be defined as $\mathtt{Attention}(Q,K,V) = \mathtt{Softmax}(\frac{QK^T}{\sqrt{d}})\cdot V$ with:
\begin{equation}
\label{eq:st_attn}
Q = W^Q z_{v_i}, K = W^K z_{\Gamma}, V = W^V z_{\Gamma},
\end{equation}
where $z_{v_i}$ denotes the feature of frame $v_i$, $z_{\Gamma}$ denotes the concatenated features of sampled reference frames $\Gamma$, the projection matrices $W^Q, W^K, W^V$ are learnable. In $\mathtt{\textcolor{cyan}{PSTConv}}$\textcolor{cyan}{\emph{-f}}, we stack a $1D$ temporal convolution with dynamic kernel size $k_\emph{f}$ after each $2D$ convolution. In the context of both $\mathtt{\textcolor{cyan}{PSTAttn}}$\textcolor{cyan}{\emph{-f}} and $\mathtt{\textcolor{cyan}{PSTConv}}$\textcolor{cyan}{\emph{-f}}, our pyramid spatial-temporal mechanism for tempo modulation is characterized by an increase in both $r_\emph{f}$ and $k_\emph{f}$ as the value $\emph{f}$ increases in the U-Net layers. This means that we progressively refine each frame's motion-aware features as by selecting more reference frames and allowing for a broader spatial-temporal correlation perception field.

\begin{figure}[t!]
\begin{center}
\includegraphics[width=1.0\linewidth]{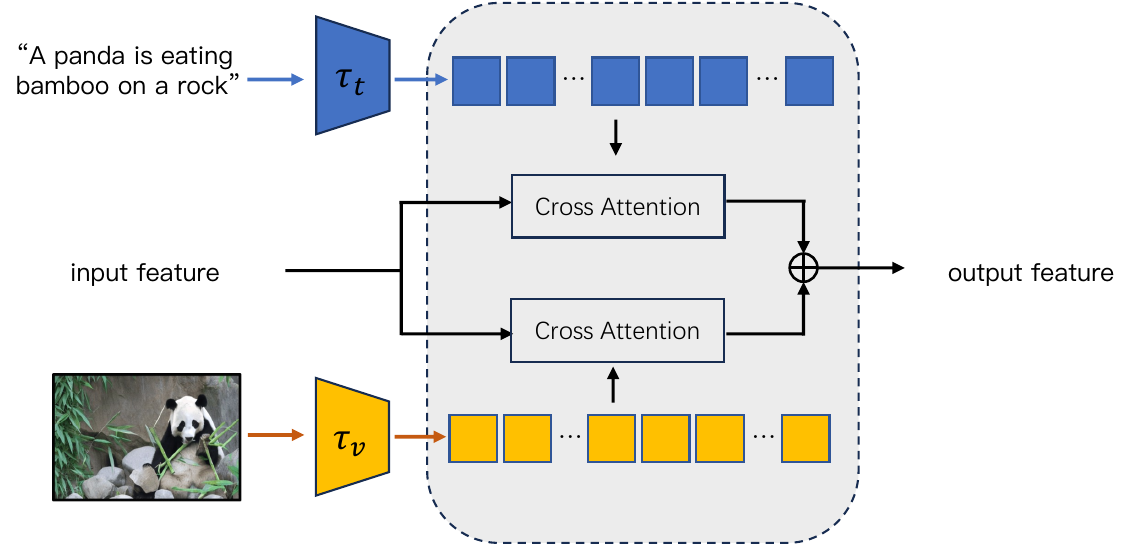}
\end{center}
\caption{An illustration of our proposed dual-stream cross-attention ($\mathtt{CrossAttn_{Dual}}$). We use the CLIP text encoder $\tau_t$ to obtain linguistic token embeddings $E_t$ for text prompt $\mathcal{C}_t$. For the image prompt $\mathcal{C}_v$, we use the counterpart CLIP image encoder $\tau_v$ to obtain the visual patch embeddings $E_v$. Then we perform a two-path cross-attention between the input feature and these bimodal embeddings. The output feature is obtained by fusing the dual-stream results with coefficient $\lambda_t$ and $\lambda_v$.}
\label{fig:crossattn}
\end{figure}

\subsubsection{Dual-Stream Cross-Attention}
While text prompt has shown to be a flexible user interface in video synthesis, how to generate a video with appearance similar to the user-provided reference image remains a problem of immense practical importance. In this paper, we aim to incorporate both controls from text descriptions and reference images into our \emph{Univid}, where the input textual and visual information complement each other to provide a detailed video description. Our core idea to introduce the visual control into the vanilla $\mathtt{CrossAttn}$ module, resulting in a dual-stream cross-attention module, namely $\mathtt{\textcolor{magenta}{CrossAttn_{Dual}}}$.

To alleviate the challenge of cross-modal alignment, we employ the paired text $\tau_t$ and image encoder $\tau_v$ from CLIP~\cite{radford2021learning} as feature extractors for the text prompt $\mathcal{C}_t$ and image prompt $\mathcal{C}_v$. Given a sequence of linguistic token embeddings $E_t = \tau_t(\mathcal{C}_t) = [e_t^1,e_t^2,...,e_t^{N_t}]$ and a sequence of visual patch embeddings $E_v = \tau_v(\mathcal{C}_v) = [e_v^1,e_v^2,...,e_v^{N_v}]$, where $N_t$ and $N_v$ are the numbers of linguistic tokens and visual patch tokens, our dual-stream cross-attention module can be formulated as:
\vspace{-1mm}
{\small
\begin{align}
\label{eq:dual_cross_attn}
z_{v_i}^\prime &= \lambda_t\; \mathtt{Attention}(Q, K_t, V_t) + \lambda_v\; \mathtt{Attention}(Q, K_v, V_v) \nonumber \\
&= \lambda_t\; \mathtt{Softmax}(\frac{QK_t^T}{\sqrt{d}})V_t + \lambda_v\; \mathtt{Softmax}(\frac{QK_v^T}{\sqrt{d}})V_v,
\end{align}}
where $Q = W^Q z_{v_i}$, $K_t = W^{K_t}E_t$, $V_t = W^{V_t} E_t$, $K_v = W^{K_v} E_v$, $V_v = W^{V_v} E_v$, $W^Q, W^{K_t}, W^{V_t}, W^{K_v}, W^{V_v}$ are the projection matrices, $z_{v_i}$ is the feature of frame $v_i$ and $z_{v_i}^\prime$ is the output feature. In contrast to a similar technique~\cite{ye2023ip} that only used the global image token, we utilize all patch tokens from the image encoder because we aim to precisely align the appearance information of each frame with the reference image in a fine-grained manner.

Since our $\mathtt{\textcolor{magenta}{CrossAttn_{Dual}}}$ align each frame's semantics with the input prompts independently, we additionally introduce a temporal attention layer $\mathtt{\textcolor{magenta}{TemAttn}}$ to propagate the semantic information across all frames. This temporal attention layer acts as a self-attention mechanism, operating specifically over the temporal axis, which is formulated as follows (using $\texttt{einops}$~\cite{rogozhnikov2021einops} notation):
\vspace{-1mm}
{\small
\begin{align}
\label{eq:temp_attn}
z' &\leftarrow \texttt{rearrange}(z, \; \texttt{b t c h w} \rightarrow \texttt{(b h w) f c}) \nonumber \\
z' &\leftarrow \mathtt{\textcolor{magenta}{TemAttn}}(z') \nonumber \\
z' &\leftarrow \texttt{rearrange}(z', \; \texttt{(b h w) f c} \rightarrow \texttt{b t c h w}),
\end{align}}
where the $\texttt{b,t,c,h,w}$ denotes the dimensions of batch, time, channel, height and weight.




\section{Experiments}
\label{sec:exp}

\begin{figure*}[ht!]
\begin{center}
\includegraphics[width=1.0\linewidth]{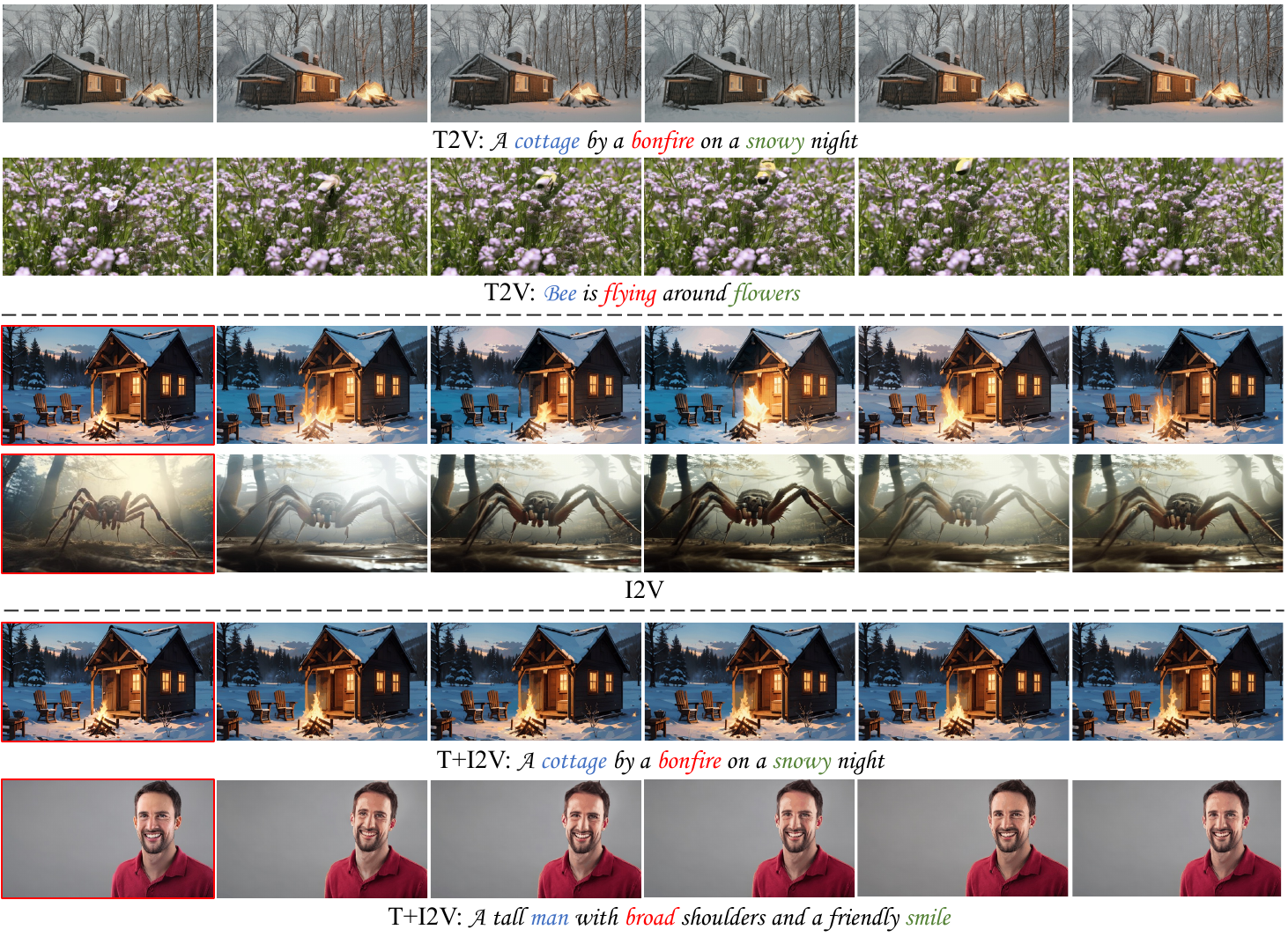}
\end{center}
\caption{The illstration displays examples generated from three tasks: T2V, I2V, and (T+I)2V. The top two rows present the T2V results, the middle two rows show the I2V results, and the final two rows illustrate the (T+I)2V results. The highlighted images in red correspond to the reference images.}
\label{fig:qualitative_pyoco}
\end{figure*}

In this section, we will present our training datasets, training strategies and implementation details. Then we both qualitatively and quantitatively evaluate our \emph{Univid} by comparing to existing approaches on the zero-shot video generation setting.

\subsection{Experimental Setups}
\noindent \textbf{Datasets} \;
To train our models, we adapt two publicly available datasets, namely Laion, and Pandas-70M. The full-version of Laion dataset contains 5 billion image-text pairs. Since our model inherits 2D parameters from SD backbone and to save training time, we only select 600 million data points with high aesthetic scores which is annotated in laion website\footnote{https://laion.ai/blog/laion-aesthetics/}. The Pandas-70M~\cite{chen2024panda} dataset contains 70 million video-text pairs. We select 6M higher-quality samples~\cite{pku_yuan_lab_and_tuzhan_ai_etc_2024_10948109} from Pandas-70M data for our model training.

\noindent \textbf{Evaluation Metrics} \;
For UCF-101, Frechet Video Distance (FVD) and Inception Score (IS)~\cite{ho2022video} are applied to assess the quality and diversity of our generate videos. For MSR-VTT, Frechet Inception Distance (FID)~\cite{parmar2022aliased}, CLIP-FID and CLIP-SIM~\cite{kynkaanniemi2023the} scores are used to evaluate our model.

\noindent \textbf{Implementation Details} \; We use the publicly available Stable Diffusion 2.1~\cite{rombach2022high} as our base T2I model and OpenCLIP~\cite{ilharco_gabriel_2021_5143773} as our image encoder $\tau_v$. In implementing a joint fine-tuning strategy that leverages both image and video data, we adopted a distinct training process for the temporal and spatial layers. For image inputs, we exclusively updated the spatial layer. In contrast, for video inputs, updates were made to both layers. 

To train our (T+I)2V model, we employ a training strategy from T2V to (T+I)2V. To accommodate the resolution of the dataset, we set the output size of the videos to 24 x 576 x 320. We first train our model by only text prompt condition on Pandas-70M and laion datasets. In this stage, we train our based T2V model which is extended from SD 2.1. Then, we integrate T2V, I2V and (T+I)2V etc. tasks to train the adapters and the corresponding cross-attention modules. Finally, we finetune the model with text and image joint conditions, update the parameters of the full U-Net backbone, allow for further alignment and joint optimization of the condition guidance between text and image.

\subsection{Main Results}

\subsubsection{Quantitative Comparisons}

\setlength{\tabcolsep}{4pt}
\renewcommand{\arraystretch}{1}

\begin{table}[t!]
\centering
\caption{Zero-shot text to video generation on UCF-101 and MSR-VTT.}
\label{tab:ucf_zeroshot}
\begin{tabular}{lcccc}
\toprule
\multirow{2}{*}{Method}   & \multicolumn{2}{c}{UCF-101} & \multicolumn{2}{c}{MSR-VTT} \\
\cline{2-5}
& IS  $(\uparrow)$   & FVD  $(\downarrow)$ & CLIP-SIM  $(\downarrow)$   & FID $(\downarrow)$  \\
\midrule
CogVideo~\cite{hong2022cogvideo} & 25.27 & 701.59 & 0.2631 & - \\
Make-A-Video~\cite{singer2022make}         & 33.00 & 367.23 & 0.3049 & - \\
MagicVideo~\cite{zhou2022magicvideo} & -& 655 & - & 36.50 \\
Latent-Shift~\cite{an2023latent} & -& 360.04&  0.2773 & - \\
Video LDM~\cite{blattmann2023videoldm}   & 33.45 & 550.61 & 0.2929 &- \\
VideoFactory~\cite{wang2023videofactory}  & - & 410.00 & 0.3005 & - \\
PYoCo~\cite{ge2023preserve}  & \textbf{47.76} &  355.19 & - & 22.14 \\
Show-1~\cite{zhang2024show} & 35.42 & 394 & \textbf{0.3072} & - \\
VideoPoet~\cite{kondratyuk2023videopoet} & 38.44 & 355 & 0.3049 & - \\
\emph{Univid} & 43.23 &  \textbf{352.77} &{0.3008}  & \textbf{22.06} \\
\bottomrule
\end{tabular}
\end{table}


\begin{table}[t!]
\centering
\caption{Zero-shot text with image and and image-only to video generation on UCF-101 and MSR-VTT.}
\label{tab:ucf_zeroshot1}
\begin{tabular}{lccccc}
\toprule
\multirow{2}{*}{Method} & \multirow{2}{*}{Task}   & \multicolumn{2}{c}{UCF-101} & \multicolumn{2}{c}{MSR-VTT} \\
\cline{3-6}
 &           & IS  $(\uparrow)$   & FVD  $(\downarrow)$  & CLIP-FID  $(\downarrow)$   & FID $(\downarrow)$ \\
\midrule
PYoCo~\cite{ge2023preserve} & T2V & \textbf{47.76} &  355.19 & 9.73 & 22.14 \\
EMU~\cite{abs-2311-10709} & I2V & 42.7 & 606.2  & - & - \\
SVD~\cite{blattmann2023stable} & I2V & - & 242 & - & - \\
DynamiCrafter~\cite{xing2023dynamicrafter} & (T+I)2V & - & 429.23 & - & - \\
\emph{Univid} & I2V & 44.51 &  {255.33} & 9.91  & 21.54 \\
\emph{Univid} & (T+I)2V &  {44.98}  &  \textbf{239.57} & \textbf{9.52} & \textbf{21.17} \\
\bottomrule
\end{tabular}
\end{table}


We quantitatively compare our method against Make-A-Video~\cite{singer2022make}, CogVideo~\cite{hong2022cogvideo}, and several concurrent works~\cite{blattmann2023videoldm,zhou2022magicvideo,xing2023dynamicrafter,wang2023videofactory,kondratyuk2023videopoet}. Table~\ref{tab:ucf_zeroshot} shows that our method outperforms all the baselines on the UCF-101 dataset and improves the zero-shot FVD from $701.59$ to $352.77$. It indicates that the videos generated by our method possess high quality, and the Inception Score (IS) has reached an advanced level, demonstrating that the diversity of our generated videos is sufficient. Meanwhile, our approach demonstrates superior performance across various metrics on the MSR-VTT dataset, particularly achieving state-of-the-art results in the FID metric~\cite{kynkaanniemi2023the}. This not only validates the effectiveness of our method in text-to-video (T2V) tasks but also indicates that the diversity of the videos generated by our model is sufficiently high.

In Table~\ref{tab:ucf_zeroshot1}, on I2V and (T+I)2V tasks, it can be seen that our baseline model achieves a new state-of-the-art FID score of $21.17$, while using ensemble models further improves both CLIP-FID and FID scores. Compared to other I2V or (T+I)2V methods, our approach demonstrates significant performance improvements in zero-shot testing on both the MSR-VTT and UCF-101 datasets. This enhancement indicates that a unified framework and training methodology not only preserves but also enhances the model's performance, leading to improved results in instruction following, generation quality, and diversity of video generation. This finding is of considerable importance.

\subsubsection{Visualized Analysis}
We present the qualitative visualizations of our \emph{Univid} on three tasks of T2V, I2V and (T+I)2V in \cref{fig:qualitative_pyoco}.
In the first case of \cref{fig:qualitative_pyoco}, our \emph{Univid} is capable of generating a coherent process of "A cottage by a bonfire on a snowy night" and keeping the irrelevant bonfire beating. In the middle two rows of \cref{fig:qualitative_pyoco}, a reference image is inputted to generate a video, and the generated video can effectively maintain the original structure of the image while exhibiting realistic motion. In the bottom two rows of \cref{fig:qualitative_pyoco}, text and image are combined to guide the generation of videos, the generated videos not only retain the informational content of the images but also effectively adhere to the motion directives provided by the text.

\section{Conclusion}
In this paper, we propose an unified video generation model (\emph{Univid}), which combines multiple tasks of T2V, I2V and (T+I)2V in a single framework. To support these tasks, a dual-stream cross-attention mechanism is designed to enable text and images to simultaneously guide video generation through re-weighted control scores. Moreover, the original spatial attention and convolution modules have been transformed into pyramid spatial-temporal modules to enhance our model's spatial-temporal modeling capability, thereby achieving improved continuous high quality video. The experimental results and visualizations indicate that our method performs well across multiple tasks, particularly excelling in maintaining semantic and temporal consistency. These demonstrate the effectiveness of our unified video generation framework.

\bibliographystyle{IEEEbib}
\bibliography{icme2025references}

\end{document}